\documentclass{article}

\usepackage{microtype}
\usepackage{graphicx}
\usepackage{subfigure}
\usepackage{booktabs} 

\usepackage{hyperref}

\usepackage[accepted]{icml2024}

\usepackage{amsmath}
\usepackage{amssymb}
\usepackage{mathtools}
\usepackage{amsthm}

\usepackage[capitalize,noabbrev]{cleveref}

\theoremstyle{plain}

\theoremstyle{definition}

\theoremstyle{remark}

\usepackage[textsize=tiny]{todonotes}

\icmltitlerunning{AI UQ to Improve Human Decision-Making}

\begin{document}

\twocolumn[
\icmltitle{Using AI Uncertainty Quantification to Improve Human Decision-Making}




\begin{icmlauthorlist}
\icmlauthor{Laura~R. Marusich}{arl}
\icmlauthor{Jonathan~Z. Bakdash}{utd}
\icmlauthor{Yan Zhou}{utd}
\icmlauthor{Murat Kantarcioglu}{utd}
\end{icmlauthorlist}

\icmlaffiliation{arl}{DEVCOM Army Research Laboratory}
\icmlaffiliation{utd}{University of Texas at Dallas, Richardson, TX}

\icmlcorrespondingauthor{Laura Cooper}{laura.m.cooper20.civ@army.mil}
\icmlcorrespondingauthor{Yan Zhou}{yan.zhou2@utdallas.edu}

\icmlkeywords{Machine Learning, ICML}

\vskip 0.3in
]



\printAffiliationsAndNotice{\icmlEqualContribution} 

\begin{abstract}
AI Uncertainty Quantification (UQ) has the potential to improve human decision-making beyond AI predictions alone by providing additional probabilistic information to users. The majority of past research on AI and human decision-making has concentrated on model explainability and interpretability, with little focus on understanding the potential impact of UQ on human decision-making.  We evaluated the impact on human decision-making for instance-level UQ, calibrated using a strict scoring rule, in two online behavioral experiments. In the first experiment, our results showed that UQ was beneficial for decision-making performance compared to only AI predictions. In the second experiment, we found UQ had generalizable benefits for decision-making across a variety of representations for probabilistic information. These results indicate that implementing high quality, instance-level UQ for AI may improve decision-making with real systems compared to AI predictions alone. 
\end{abstract}

\section{Introduction}
Using AI to improve human decision-making requires effective human-AI interaction. Recent work on human-AI interaction guidelines focuses on explainability and interpretability~\cite{amershi2019guidelines}, which may improve subjective human ratings of trust in and usability of AI. However, a quantitative synthesis of studies found that explanations may not generally improve decision accuracy beyond AI prediction alone~\cite{schemmer2022meta} in many application domains. One less-explored possibility for promoting effective human-AI interaction is AI Uncertainty Quantification (UQ) for predictions. AI UQ is posited to be key for human decision-making~\cite{abdar2021review, jalaian2019uncertain}. However, there is conflicting evidence in the existing literature as to whether presenting AI UQ for predictions can improve human decision-making accuracy, and how to best communicate this uncertainty information \cite{lai2021towards}. These conflicting results may be due in part to "a lack of discussion on the reliability of uncertainty estimates, sometimes referred to as calibration"~\citep[p. 15]{lai2021towards}.

In order to resolve these questions, we use well-calibrated, instance-level AI Uncertainty Quantification (UQ) evaluated using a strict scoring rule~\cite{doi:10.1198/016214506000001437} using the ground truth for class labels \footnote{We wish to highlight that for the classification task, ground truths for class labels are utilized to offer well-calibrated, high quality, instance-level uncertainty quantification for human subject experiments.}. We evaluate the impact of this AI UQ in two pre-registered, large sample size, online behavioral experiments assessing human decision-making. Decision-making is measured objectively using response accuracy and confidence calibration with accuracy. We found that providing high-quality AI UQ meaningfully improves decision-accuracy and confidence calibration over an AI prediction alone. Additionally, the benefits of this AI UQ appear to be generalizable -- decision-making was similar for AI UQ presented with different visualizations and types of information. Our results indicate well-calibrated AI UQ is beneficial for decision-making. 

The paper is structured as follows. In section~\ref{sec:background}, we provide the background information on uncertainty and human decision-making and an overview of existing techniques for AI UQ. Section~\ref{sec:work} describes our UQ technique and experimental design. In section~\ref{sec:exp1}, we report findings from the behavioral experiments comparing human decision making accuracy with or without UQ information. In section~\ref{sec:exp2}, we report the impact of different visualizations of UQ information. Finally, sections~\ref{sec:gendisc} and~\ref{sec:conc} conclude by discussing the implications of our results and future work. 

\section{Background and Related Work}
\label{sec:background}
\subsection{Human Decision-Making and Uncertainty}
The possible benefit of AI UQ is supported by work in the judgment and decision-making literature on decision-making under uncertainty. This work shows that providing overall prediction uncertainty enhances decision-making accuracy. For example, in weather forecasting, humans demonstrate higher decision-making performance when they receive well-calibrated probabilistic information (e.g., a forecast with a probability of rain), compared to only deterministic predictions (e.g., it will or will not rain)~\cite{frick2011can, joslyn2013decisions, morss2008communicating, nadav2009uncertainty}. However, increasing information, even when it is task-relevant, is not always beneficial to human decision-making performance \citep[e.g.,][]{marusich_info, gigerenzer2009homo, Alufaisan_Marusich_Bakdash_Zhou_Kantarcioglu_2021}. An additional consideration is the way that uncertainty information is represented. In human decision-making, communicating uncertainty with visual representations and other intuitive methods can be especially effective~\cite{gigerenzer2007helping, hullman2018pursuit}. 

Despite previous general findings that uncertainty information is useful for decision-making, there is limited behavioral research assessing the benefits of AI UQ, particularly for human decision-making accuracy. Among studies that do assess objective accuracy performance \citep[e.g.,][]{zhang2020effect, bucinca2021}, both the methods and results vary. In particular, the quality of the UQ calibration varies, with some studies opting to simulate AI prediction confidence with wizard-of-oz techniques, and others using the prediction probabilities generated by their model, but without quantifying the calibration of those probabilities. As a result, the potential benefits for AI UQ remain at least somewhat of an open question \cite{lai2021towards}.

There is a clear gap for behavioral studies assessing human decision-making performance using quantifiably well-calibrated AI UQ for predictions. Our method for AI UQ uses known class labels to ensure high-quality uncertainty information at the instance-level, as poorly calibrated uncertainty information is likely to be detrimental to decision-making. 
We emphasize that \emph{the application of known class labels to generate instance-level UQ aims to provide well-calibrated AI UQ for individual predictions specifically in the context of human subject experiments}. This approach is not \emph{designed for real-life deployment scenarios where class labels may not be known in advance}.
In the next section, we briefly provide context of existing techniques for AI UQ, which are often model-based and typically do not require labelled data.

\subsection{Techniques for AI UQ}
Predictions by AI-based systems are subject to uncertainty from different sources. The source of uncertainty is either aleatoric, caused by noise in data and irreducible, or epistemic because of uncertain model distribution~\cite{NIPS2017_2650d608}. Uncertainty quantification methods have been developed to assess the reliability of AI predictions~\cite{10.1016/j.inffus.2021.05.008}, including Bayesian methods and ensemble methods~\cite{abdar2021review}.  

Monte Carlo sampling~\cite{neal2012bayesian} and Markov chain Monte Carlo ~\cite{10.1145/1390156.1390267,pmlr-v37-salimans15,pmlr-v32-cheni14,NIPS2014_21fe5b8b,10.5555/2969442.2969494,2a464d582f3c4c4c8ceac529e8088282,NEURIPS2019_c055dcc7} are heavily used for uncertainty quantification in Bayesian techniques~\cite{NIPS2017_2650d608,2cb6ee3631cc4275aedc26a2124c1d7d,DBLP:conf/iccv/0009JLZ0WH19}. To estimate aleatoric uncertainty, a hidden variable is often proposed to represent the underlying data point $x^*$ from which a given instance $x$ is only one of many possible observations of $x^*$. Parameters modeling the transformation from $x^*$ to $x$ can be sampled to obtain multiple copies of the hidden $x^*$. For epistemic uncertainty, the distribution of model parameter $\theta$ is often approximated during training by achieving certain objective optimization, for example, the Kullback–Leibler divergence. The distribution of the prediction can be sampled from the samples of the learned model parameters. The predictive uncertainty can be established from the variance or entropy of the sampled predictions of the sampled hidden states of a given instance. 

Quantifying uncertainty on learning models from a Bayesian perspective takes many different forms. Uncertainty Posterior distribution over BNN weights can be learned using variational inference~\cite{Subedar_2019_ICCV,10.5555/3305890.3305910,DBLP:journals/corr/abs-2002-03704,DBLP:conf/iclr/GhoshSVBS20}. On the other hand, Generative Adversarial Networks (GANs) are used to generate out-of-distribution (OoD) examples~\cite{oberdiek2022uqgan}.  Implicit neural representations (INRs) are reformulated from a Bayesian perspective to allow for uncertainty quantification~\cite{vasconcelos2023uncertainr}. Similarly, Direct Epistemic Uncertainty Prediction (DEUP) is proposed to address the issue that using the variance of the Bayesian posterior does not capture the epistemic uncertainty induced by model misspecification~\cite{lahlou2023deup}. Aleatoric uncertainty and epistemic uncertainty have also been modeled as universal adversarial perturbations~\cite{DBLP:conf/iccv/0009JLZ0WH19}.

Ensemble models can enhance the predictive accuracy, however, it is highly debated whether an ensemble of models can provide a good uncertainty estimate~\cite{abdar2021review,10.5555/3495724.3496118,10.5555/3327144.3327239}. Recently, benefits of prior functions and bootstrapping in training ensembles with estimate of uncertainty have been discussed~\cite{dwaracherla2023ensembles}. Maximizing Overall Diversity takes into account ensemble predictions for possible future input when estimating uncertainty~\cite{Jain_Liu_Mueller_Gifford_2020}. Random parameter initialization and data shuffling have also been proposed to estimate the uncertainty of DNN ensembles~\cite{10.5555/3295222.3295387}. A Bayesian non-parametric ensemble (BNE) approach is proposed to account for different sources of model uncertainty~\cite{NEURIPS2019_1cc8a8ea}.
More details on extensive studies on quantifying uncertainty with respect to both Bayesian and ensemble methods, as well as in real applications can be found in~\cite{abdar2021review}. 

However, prediction probabilities are prone to overconfidence in some AI models. There is a lack of discussion on the calibration of uncertainty estimates in the existing literature. 

In this work, we achieved efficiency of UQ estimate by assessing the change of prediction yielded from repeatedly sampling noise adjacent to a given instance, and carefully calibrated the uncertainty information shown to the user by leveraging the ground truth. 
More precisely, \emph{we provide well-calibrated uncertainty estimates in different visualizations of confidence intervals to the human participants}. 
Unlike the existing work discussed above, our goal is to \textit{provide the uncertainty information to the human participants} to understand whether well-calibrated \textit{uncertainty quantification information helps in user decision-making}. To achieve this goal, we do not attempt to come up with a UQ method a priori. Instead, we take the liberty of knowing the true labels of given instances, and simplify the problem as sampling predictive confidence from instances distorted with a small amount of random noise. The quality of the disclosed uncertainty estimate is verified using a strictly proper scoring rule~\cite{doi:10.1198/016214506000001437} prior to use in two behavioral experiments. While there have been recent calls for research using UQ with human decision-making (e.g., \citealt{bhatt_review, lai2021towards}), the few existing studies tend to focus on qualitative or subjective assessments of human behavior (e.g.,~\citealt{Prabhudesai_qual}). Furthermore, it is not clear how useful to decision-makers the UQ information provided in these studies is, \textit{due to lack of proper calibration}.

\section{Current Work}
\label{sec:work}
We conducted two experiments to assess the effect of providing visualizations of AI prediction UQ information upon the accuracy and confidence of human decision-making. The first experiment compares performance when AI uncertainty is provided to performance when only an AI prediction, or no AI information at all, is provided. The second experiment compares decision-making performance for different representations of AI uncertainty.

Our methods and results for the instance-level predictive UQ and behavioral experiments are fully reproducible. See the supplementary material for details and links.

In both experiments, we assessed our research questions using three different publicly-available and widely-used datasets: the {\em Census}, {\em German Credit}, and {\em Student Performance} datasets from the UCI Machine Learning Repository~\cite{Dua:2019}, described in more detail below.

\subsection{Datasets}
The {\em Census} dataset has 48,842 instances and 14 attributes. The missing values in the dataset were replaced with the mode (the most frequent value), and the dollar amounts were adjusted for inflation. The {\em German Credit} dataset has 1,000 instances and 20 attributes. The currency values were converted to dollars and adjusted for inflation. The {\em Student Performance} has 649 instances and 33 attributes. Three of the attributes {\em first period grade}, {\em second period grade}, and {\em final grade} were combined into one with their average. Each dataset was split into training (70\%) and test (30\%) data sets. 

We selected these datasets because they involve real-world contexts that are fairly intuitive for non-expert human participants to reason about (e.g., will a student pass or fail a class?). In addition, using three datasets that vary in number of features and in the overall accuracy classifiers can achieve in their predictions ensures that our findings are not limited only to one specific dataset.

Several machine learning models were trained on all three datasets, including decision tree, logistic regression, random forest, and support vector machine. The best set of hyper-parameters was determined through grid search. Random forest was the best in terms of overall accuracy on the datasets and therefore was selected for use as the AI model in this study. The mean accuracy on the {\em Census} data is 85.3\%, 75.7\% on the {\em German Credit} data, and 85.1\% on the {\em Student Performance} data. All classification tasks were completed on an Intel\textsuperscript{\textregistered} Xeon\textsuperscript{\textregistered} machine with a 2.30GHz CPU. 

\begin{figure*}[h]
\centering
\begin{minipage}{0.335\textwidth}
\includegraphics[width=\textwidth]{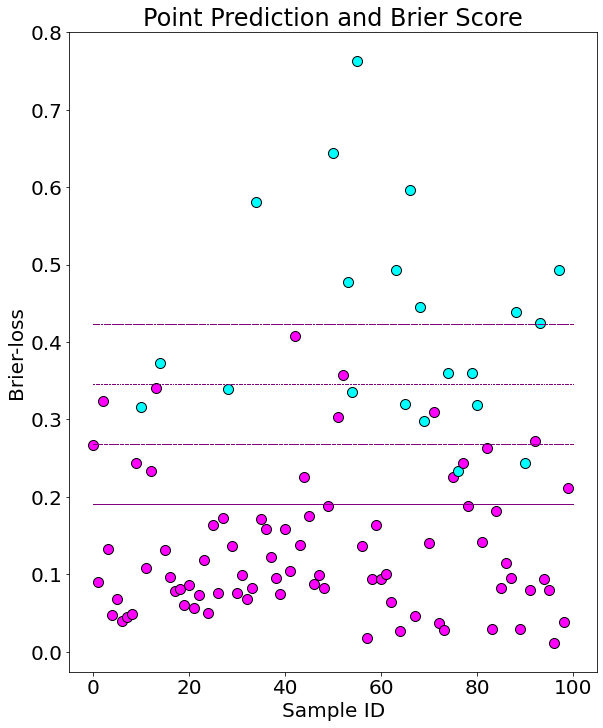}
\centering\mbox{Census}
\end{minipage}
\begin{minipage}{0.32\textwidth}
\includegraphics[width=\textwidth]{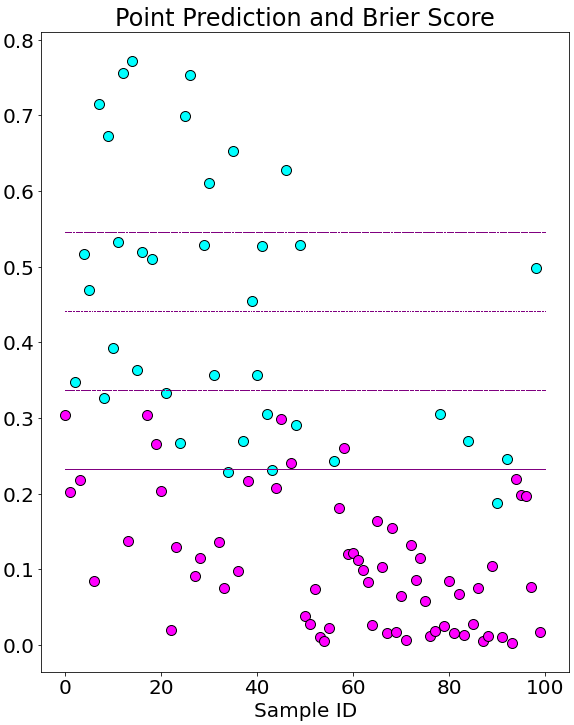}
\centering\mbox{German Credit}
\end{minipage}
\begin{minipage}{0.32\textwidth}
\includegraphics[width=\textwidth]{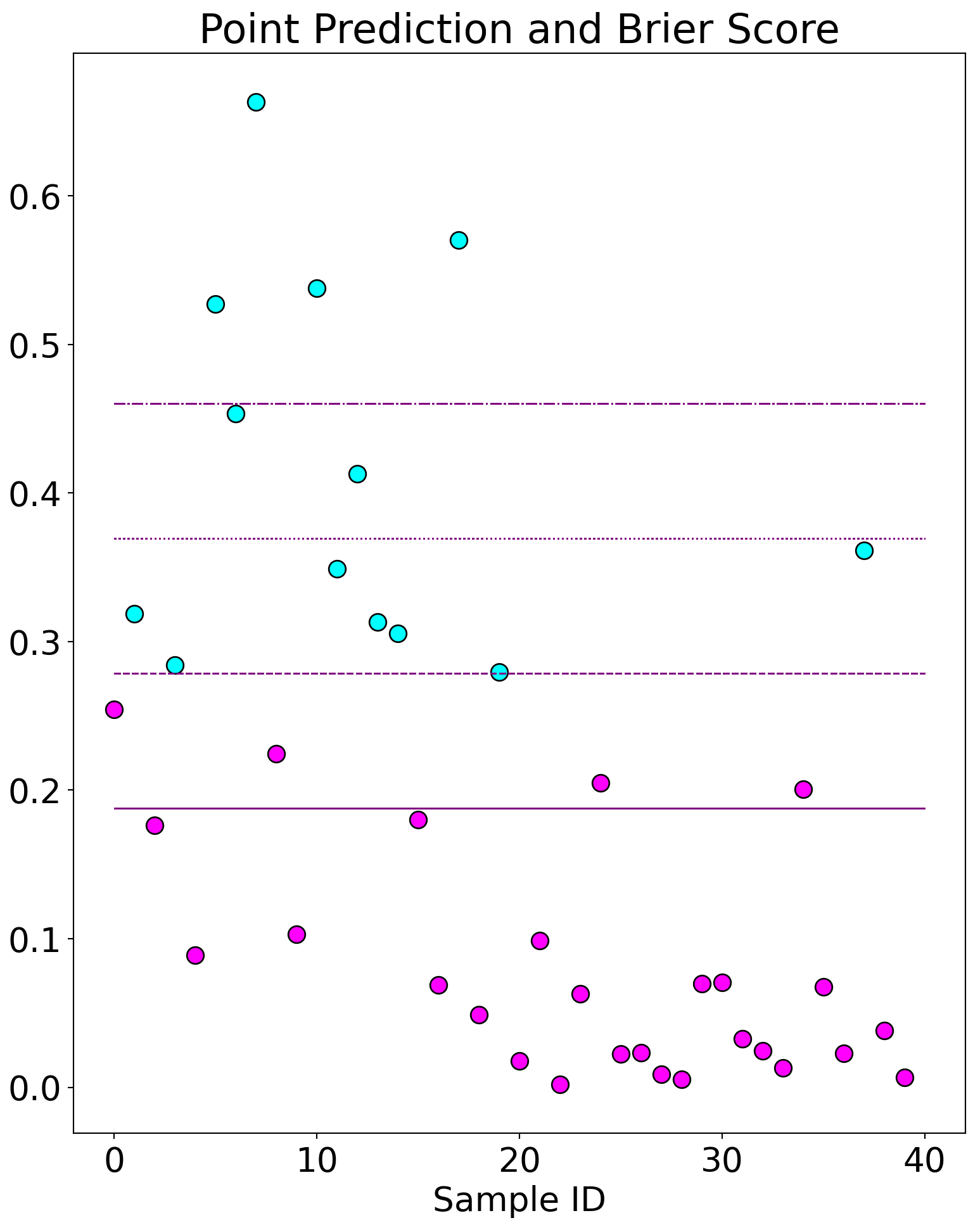}
\centering\mbox{Student Performance}
\end{minipage}
\caption{\label{fig:brier}The Brier score of the ``cloned'' instances for {\em Census} (100), {\em German Credit} (100), and {\em Student Performance} (40) sampled for demonstration. $Y$-axis is the Brier score. Magenta marks the samples that are correctly predicted by the AI model, cyan marks samples incorrectly predicted by the model. Horizontal lines illustrate the mean of the Brier score, and its 0.5, 1, and 1.5 standard deviations.}
\end{figure*}

\subsection{Instance-Level Predictive Uncertainty Quantification}
UQ methods in existing literature estimate predictive uncertainty without the knowledge of the true labels of the test instances. These methods are subject to complicated calculations, sometimes poor convergence, lack of scalability, and sometimes, they are time and resource consuming~\cite{10.1016/j.inffus.2021.05.008}. In our study, we aim to provide predictive uncertainty quantification to human decision-makers and use the advantage of knowing the true labels in advance. Therefore, we simplify the problem as sampling predictive confidence from samples of $x$ with a small random disturbance and \textit{verify the quality of the uncertainty estimate} using a strictly proper scoring rule~\cite{doi:10.1198/016214506000001437} before showing it to the human. Note that, without knowing the ground truth, this treatment of UQ would be reckless and naive. It would appear that we model a prior distribution over hypothesis as the distribution over observations in the noisy neighborhood of a given instance. However, given the true label of an instance, we can hypothesize that observations over its $n$ neighboring noisy samples are $n$ plausible fits for this instance, and confirm our hypothesis with a strictly proper scoring rule. 

Predictive uncertainty consists of data uncertainty (aleatoric) and model uncertainty (epistemic). To model data uncertainty, we sample $n$ instances from a Gaussian distribution within a standard deviation $\sigma$ from a given instance $x$, assuming $x = x^* + \eta$ where $x^*$ is the clean input of $x$ without the random disturbance $\eta$. Thus, given a prediction function parameterized by $w$, the class label of $x$ is predicted as:
\[
p(y|x, w) = \int p(y|x^*, w) p(x^*|x) dx^*
\] 
The posterior $p (x^* | x)$ is generally unknown. By assuming $\eta \sim \mathcal{N} (0, \sigma_0^2,I)$, we can sample from the posterior distribution given the noisy input $x$. In this study, we set $n = 100$ and $\sigma_0 = 0.1$.

Similarly, for model uncertainty, given a set of training data $(X,Y)$, we assume there exists an uncertain set of $m$ models with model uncertainty $\theta^{(m)} \sim p(\theta |X,Y)$. Hence, given an instance $x$, the probability of the class label of $x$ is:
\[
p(y|x, X,Y) = \mathbb{E}_{p(\theta|X,Y)}[p(y|x,\theta)].
\]

In this study, we tested an ensemble of {\em logistic regression}, {\em support vector machine}, and {\em random forest} to predict the class label. The best uncertainty estimate, however, was obtained by using the {\em random forest} alone, assessed by the Brier score discussed below.  

Predictive uncertainty per instance was computed for 294 randomly selected {\em Census} instances, 300 {\em German Credit} instances, and 194 {\em Student Performance} instances, for use in the behavioral study.
Predictive uncertainty at the instance-level was measured on random samples in the neighborhood of the instance. More specifically, given an instance $x$, $n$ random ``clones'' were sampled from a Gaussian distribution within $\delta$ standard deviation from the mean $x$. In the experiment, we let $n=100$ and $\delta=0.1$ which provided sufficient statistical significance and constrained neighborhood choices. Class probabilities were computed using the trained random forest classifier for each of the 100 samples, and the 95\% confidence interval of the class probabilities was used as the predictive uncertainty range for instance $x$. UQ computed from {\em random forest} alone was superior to that of the ensemble of {\em logistic regression}, {\em support vector machine}, and {\em random forest}, hence was used in the behavioral study.       

Knowing the ground truth (class label) of the instances, we can verify the quality of the simulated predictive uncertainty using the Brier score (also referred to as Brier loss). The Brier score measures the mean squared difference between the predicted probability and the true outcome. For each selected instance $x$, with $y \in \{0, 1\}$ and the predicted probability $p_i = Pr(y_i =1)$ for each ``cloned'' sample $x_i$, we compute the Brier score $B=\frac{1}{n} \sum_{i} (y - p_i)^2 $   
between the predicted probability of the ``cloned'' samples and $y$---the actual label of $x$. 

If the ``cloned'' samples are truly representative of $x$, the computed Brier score should reflect the correctness of the prediction made for $x$ by the AI model $M$. A smaller Brier score means more accurate predictions made for the clones of $x$, and therefore should correspond to a correct classification for $x$ by $M$. We verified empirically that the Brier scores of the predictive uncertainty is highly correlated with the true prediction for $x$ by $M$, as shown in Figure~\ref{fig:brier}. Points with low Brier score corresponds to instances where $M$ is correct. In Figure~\ref{fig:brier}, points in magenta are the samples correctly predicted by the AI model, and points in cyan are samples incorrectly predicted by the model. As can be seen, ``clones'' for each correctly predicted sample correspond to low Brier score loss, and vice versa, cloned samples for incorrectly classified samples produce high Brier scores. Horizontal lines illustrate the mean of the Brier score, and its 0.5, 1, and 1.5 standard deviations. The Brier score close to mean (approximately $0.25$) is a highly accurate indicator of the classification outcome. In essence, the Brier score resembles the trust score~\cite{10.5555/3327345.3327458} that has high precision at identifying correctly classified examples, and is adequate to assess the quality of the estimated UQ.

\subsection{Behavioral Experiments: General Methods}
\begin{figure}[h]
\begin{center}
\centerline{\includegraphics[width=\columnwidth]{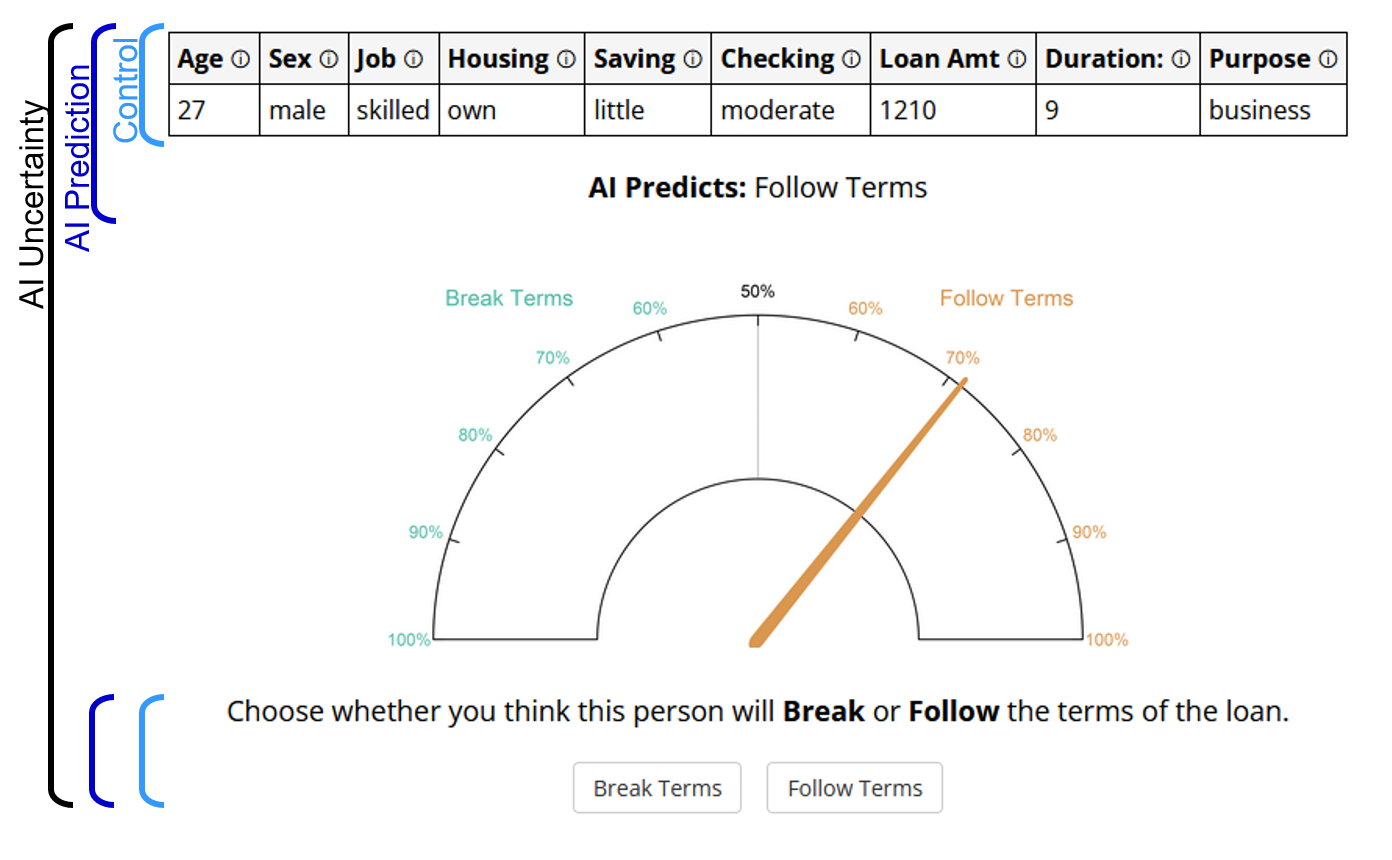}}
\caption{Example showing the information appearing in the three AI conditions in Experiment 1 for a trial from the \emph{German Credit} dataset condition.}
\label{fig:trial_example}
\end{center}
\vskip -0.5in
\end{figure}

We used the same experimental task across both Experiment 1 and 2, which was developed using jsPysch \cite{de2015jspsych} and hosted on MindProbe \url{https://mindprobe.eu/} using Just Another Tool for Online Studies (JATOS) \url{https://github.com/JATOS/JATOS}. Each trial of this task included a description of an individual and a two-alternative forced choice for the classification of that individual. Each choice was correct on 50\% of the trials, thus chance performance for human decision-making accuracy was 50\%. In some conditions, an AI prediction or an AI prediction and a visualization of prediction uncertainty would also appear. Figure \ref{fig:trial_example} shows an example of the information appearing in the three AI conditions for a trial from the German Credit dataset condition (see supplementary material for more example trials). After making a decision, participants then entered their confidence in that choice, on a Likert scale of 1 (No Confidence) to 5 (Full Confidence). Feedback was then displayed, indicating whether or not the previous choice was correct.

For each dataset, we selected 50 instances with representative average AI prediction accuracies (\emph{Census}: 88\%, \emph{German Credit}: 76\%, \emph{Student Performance}: 82\%). Then, for each participant, we randomly sampled 40 of those 50 instances for the block of test trials, resulting in small variations in AI accuracy for each participant.

Online participants were recruited from Prolific (\url{https://www.prolific.co}). They provided informed consent, viewed a series of instructional screens with examples, completed 8 practice trials, followed by 40 test trials, and a brief series of questionnaires (demographics, self-reported strategies, subjective usability, subjective task difficulty, task understanding, and an assessment of risk literacy~\cite{cokely2012measuring}, see supplementary material). Most participants completed the task in less than 20 minutes, and they were paid \$5.00 for their participation (i.e., well above the U.S. federal minimum hourly wage). This research received Institutional Review Board (IRB) approval.

\section{Experiment 1}\label{sec:exp1}
Experiment 1 compared participant decision-making accuracy in three conditions: Control (no AI prediction information), AI Prediction, and AI Uncertainty (AI prediction plus a visualized point estimate of AI uncertainty), see Figure \ref{fig:trial_example}. All hypotheses and methods were pre-registered (\url{https://aspredicted.org/ZW9_Z54}).

We hypothesized that for all three datasets, participant decision accuracy would be highest in the AI Uncertainty condition, followed by the AI Prediction condition, and lowest in the Control condition. Similarly, we hypothesized that confidence calibration (positive association between confidence and accuracy) would be strongest in the AI Uncertainty condition, followed by the AI Prediction condition, and lowest in the Control condition.
   
\subsection{Participants}
We recruited nearly 50 participants in each of 9 experimental conditions, for a total of 445 participants (48.8\% male, 48.5\% female, 2.7\% other or prefer not to answer). The majority (68.8\%) of participants were 18-44 years old. 

\subsection{Results and Discussion}

\begin{figure}[h]
\begin{center}
\centerline{\includegraphics[width=\columnwidth]{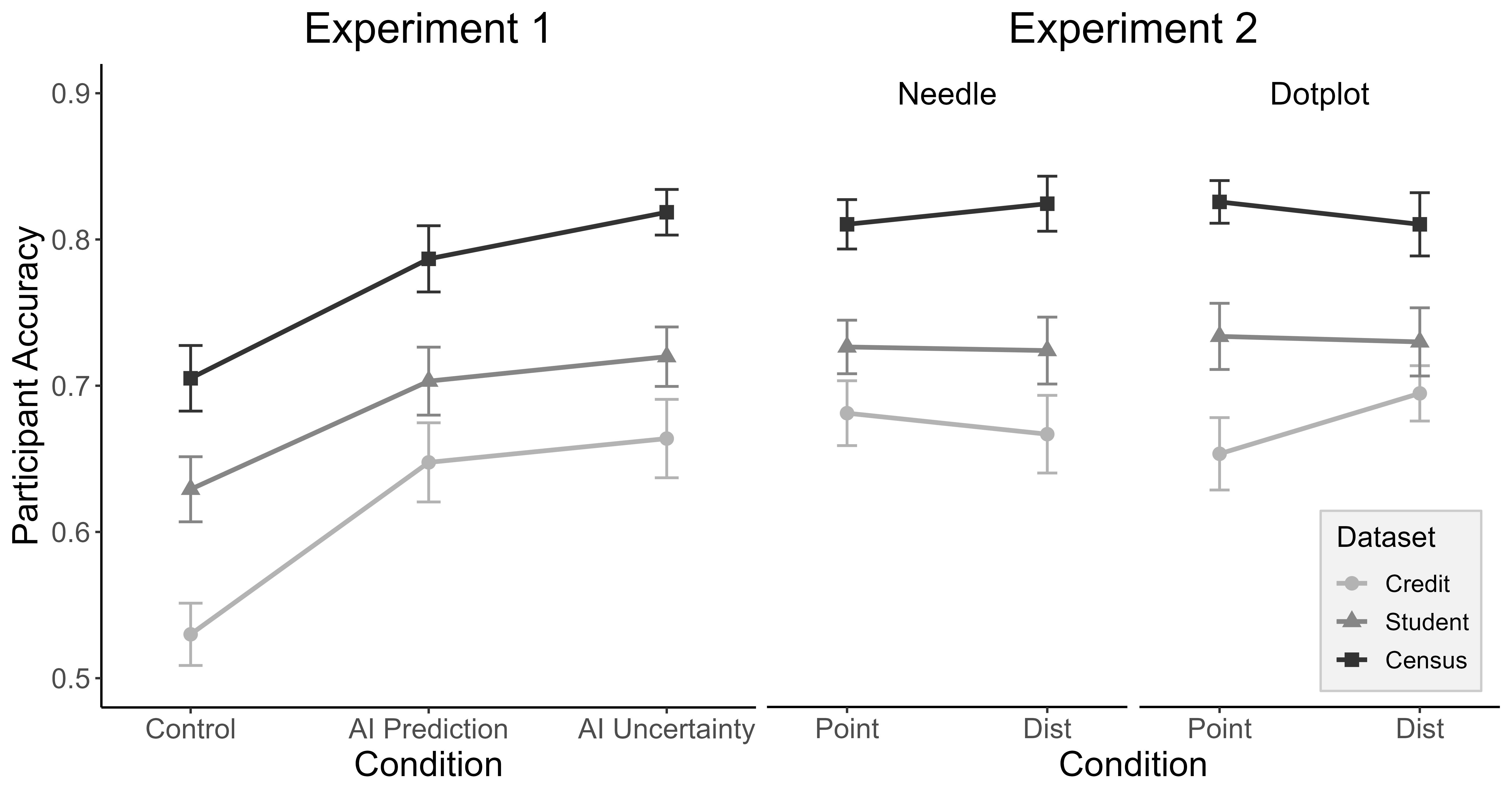}}
    \caption{
        Participant accuracy in Experiments 1 (left) and 2 (right). Error bars represent 95\% confidence intervals.}
        \label{fig:all_acc_results}
\end{center}
\vskip -0.2in
\end{figure}

We excluded trials with reaction times that exceeded three standard deviations above the mean; this resulted in the removal of 396 out of 17,800 trials across all participants. An omnibus 3 (AI Condition) x 3 (Dataset) ANOVA for mean accuracy indicated a significant main effect of AI condition ($F(2,436) = 84.11, {p} < 0.001, \eta_p^2 = 0.28$; see left side of Figure \ref{fig:all_acc_results}). This large effect size is driven primarily by the differences between the Control condition and the other two conditions. However, we also used Tukey's honest significance test to conduct post-hoc comparisons between individual conditions. These comparisons showed not only that accuracy in the AI Prediction condition was higher than in the Control condition ($t(436) = 9.91, {p} < 0.0001$), but also that accuracy in the AI Uncertainty condition was further improved (although to a lesser extent) over the AI Prediction condition ($t(436) = 2.36, {p} = 0.049$). 

There was also a significant main effect of dataset upon accuracy ($F(2,436) = 144.72, {p} < 0.001, \eta_p^2 = 0.40$). Unsurprisingly, human accuracy was lowest in the \emph{German Credit} dataset, for which AI accuracy is also relatively low, and human accuracy was highest in the \emph{Census} dataset, where AI accuracy is also relatively high.

\begin{figure}[t]
\begin{center}
\centerline{\includegraphics[width=\columnwidth]{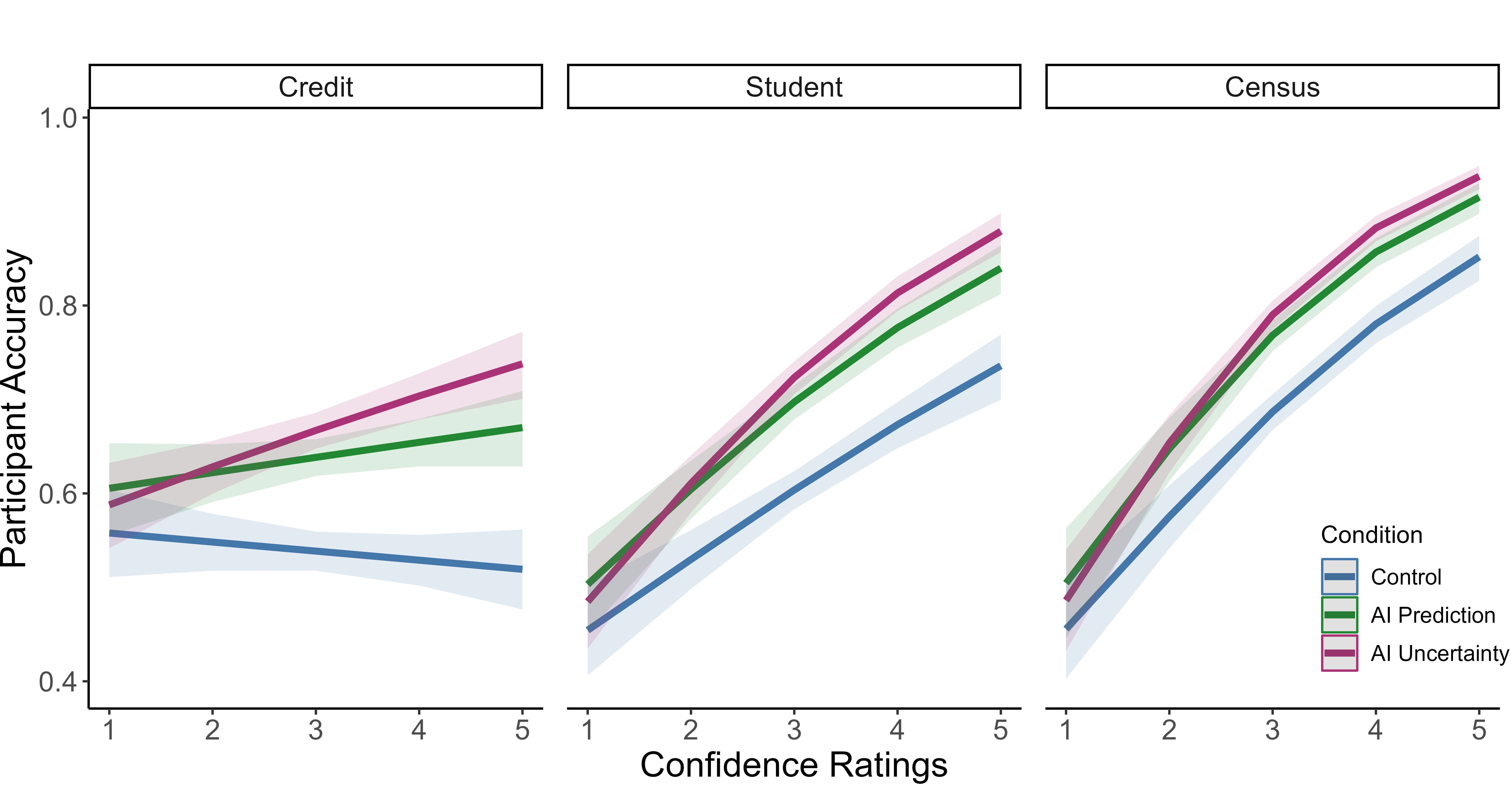}}
\caption{Predicted effects (level 2/overall results of multilevel model) of confidence ratings, dataset, and AI condition upon accuracy in Experiment 1. Steeper, positively-sloped lines indicate better confidence calibration. Shaded areas represent 95\% confidence intervals for the predicted values.}
\label{fig:Exp1_accmodel}
\end{center}
\vskip -0.2in
\end{figure}

To assess confidence calibration, we fit a multilevel model~\cite{gelman2006data} that included dataset, AI condition, and confidence ratings as fixed-effect predictors, with varying intercepts for each participant. The multilevel model accounts for a moderate amount of variance in fixed and varying effects, conditional Pseudo-$R^2$ = 0.10. This model indicated that the relationship between confidence and accuracy interacted with both dataset and AI condition, illustrated in Figure \ref{fig:Exp1_accmodel}. 

As hypothesized, confidence was most highly calibrated with accuracy in the AI Uncertainty condition, followed by the AI Prediction condition, and lowest in the Control condition. See data and code links in the supplementary material for details of the model.

\begin{figure}[!t]
\begin{center}
\centerline{\includegraphics[width=\columnwidth]{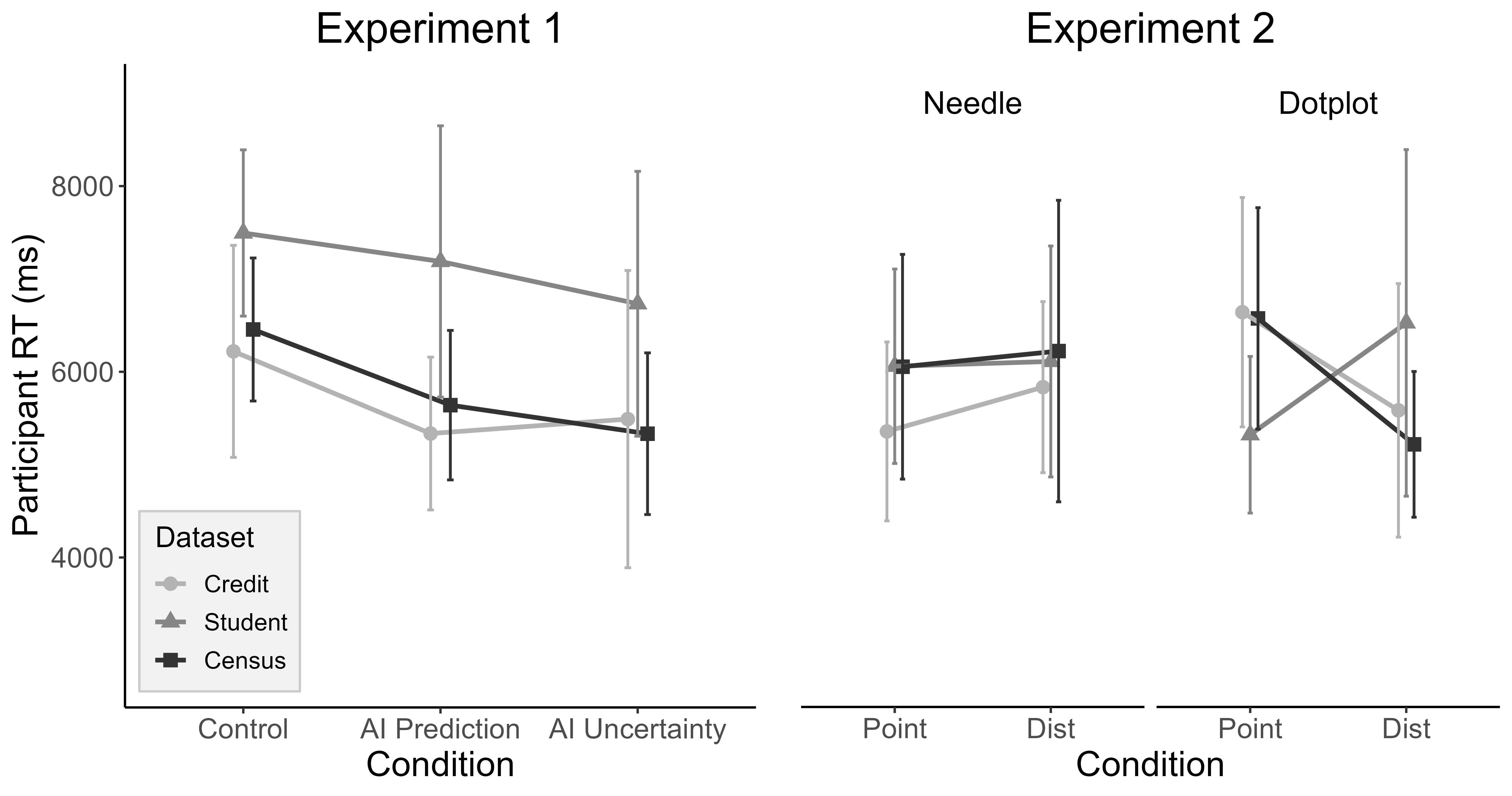}}
\caption{Participant response times (RT) in milliseconds in Experiment 1 (left) and Experiment 2 (right). Error bars represent 95\% confidence intervals.}
\label{fig:all_rt_results}
\end{center}
\vskip -0.2in
\end{figure}

We also analyzed the impact of AI condition and dataset upon participants' response times (RT). Using an omnibus 3 x 3 ANOVA, we found a significant main effect of dataset upon RT ($F(2,436) = 6.22, {p} = 0.002, \eta_p^2 = 0.03$), with participants responding slowest in the \emph{Student Performance} dataset (see left side of Figure \ref{fig:all_rt_results}), perhaps due to the larger number of attributes to consider for each instance. We did not find significant effects of AI condition ($F(2,436) = 1.97, {p} = 0.14, \eta_p^2 = 0.009$) or an interaction effect ($F(4,436) = 0.13, {p} = 0.97, \eta_p^2 = 0.001$). These results imply that the accuracy benefit for AI UQ information is not merely due to a speed/accuracy tradeoff among participants~\cite{wickelgren1977speed}.

\section{Experiment 2}
\label{sec:exp2}
\begin{figure*}[h]
\begin{center}
\centerline{\includegraphics[width=\textwidth]{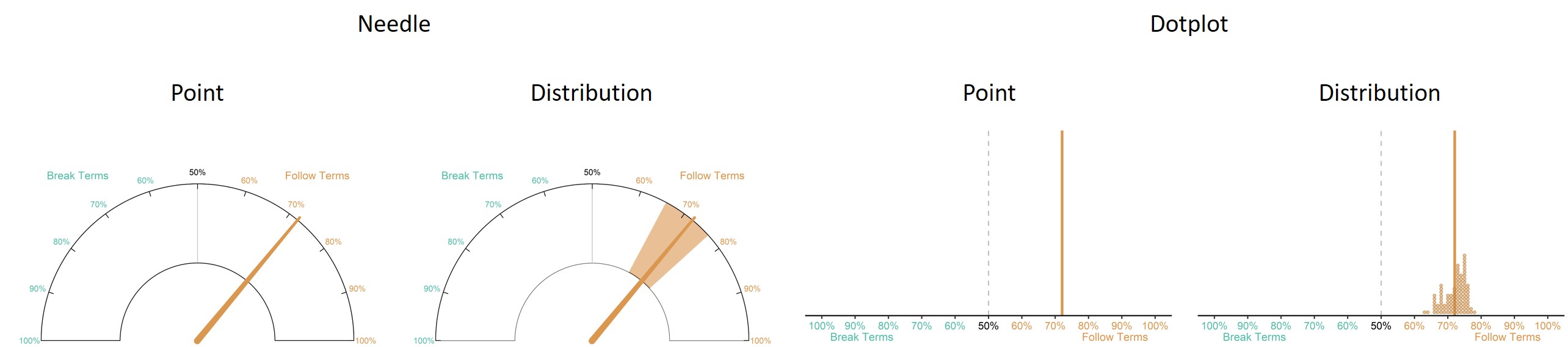}}
\caption{Example of the four conditions (point vs. distribution, and needle vs. dotplot) in Experiment 2, using a trial from the \emph{German Credit} dataset.}
\label{fig:4viz}
\end{center}
\vskip -0.2in 
\end{figure*}

Experiment 1 demonstrated that decision-making performance can be improved with AI UQ information; we designed Experiment 2 to test if different representations of UQ might be more or less beneficial for decision-making. We compared performance with distributions of uncertainty probabilities to point-estimated probabilities, as well as two different visualizations of uncertainty (needle vs. dotplot), again using the same three datasets used in Experiment 1 (see Figure \ref{fig:4viz}). All hypotheses and methods were pre-registered (\url{https://aspredicted.org/CJW_71H}).

We hypothesized that for all three datasets, both participant decision accuracy and confidence calibration would be higher with distribution information than for point-estimated UQ in the AI Uncertainty condition, followed by the AI Prediction condition, and lowest in the Control condition. We also hypothesized that, within the distribution conditions, accuracy and confidence calibration would be higher for the dotplot visualization than for the needle visualization, due to the more detailed information about the shape of distributions available in the dotplot, compared to the needle which only shows the distributions as a uniform range.

\subsection{Participants}
We recruited 50 participants in each cell, for a total of 600 participants (48.5\% male, 49.3\% female, 2.2\% other or prefer not to answer), from the Prolific platform. Most participants (75.3\%) were 18-44 years old. 


\subsection{Results and Discussion}
We excluded 553 trials (out of 24,000 across all participants) with reaction times that exceeded three standard deviations above the mean. An omnibus 2 (point vs. distribution) x 2 (needle vs. dotplot) x 3 (dataset) ANOVA for mean accuracy indicated a significant main effect of dataset ($F(2,588) = 188.77, {p} < 0.001, \eta_p^2 = 0.39$), where performance was again highest for the \emph{Census} dataset, followed by \emph{Student Performance}, and lowest for \emph{German Credit}. However, there was no significant main effects of point vs. distribution ($F(1,588) = 0.28, {p} = 0.60, \eta_p^2 < 0.001$) or visualization type ($F(1,588) = 0.16, {p} = 0.69, \eta_p^2 < 0.001$). Neither was there evidence of significant first-order interaction effects among the three manipulated variables, see right side of Figure \ref{fig:all_acc_results}).

\begin{figure}[t]
\begin{center}
\centerline{\includegraphics[width=\columnwidth]{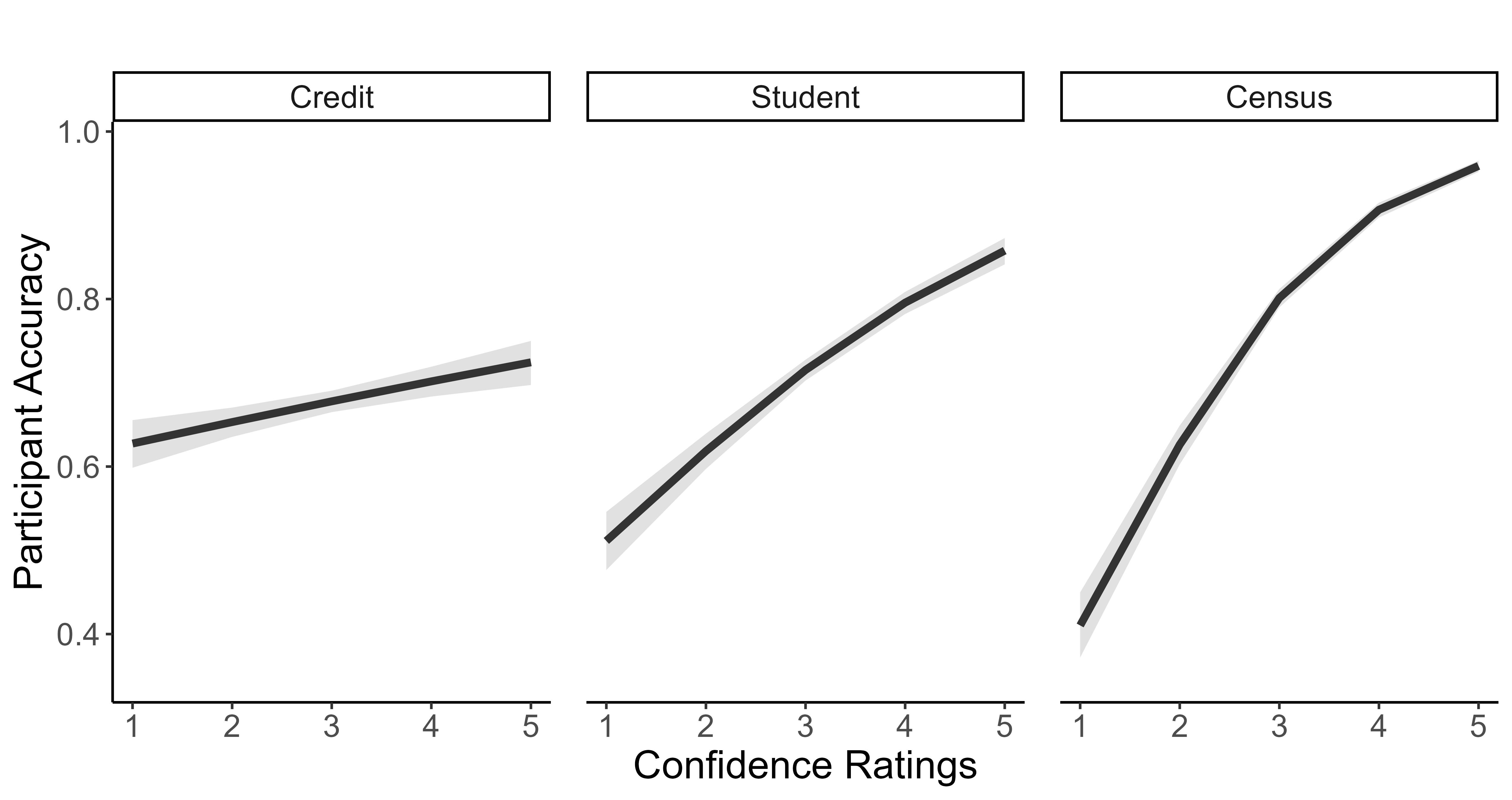}}
\caption{Predicted effects (level 2/overall results of multilevel model) of confidence ratings and dataset upon accuracy in Experiment 2. Steeper, positively-sloped lines indicate better confidence calibration. Shaded areas represent 95\% confidence intervals for the predicted values.}
\label{fig:Exp2_accmodel}
\end{center}
\vskip -0.5in 
\end{figure}

As in Experiment 1, we fit multilevel models with varying intercepts for each participant to assess confidence calibration. We found that the best-fitting model (as assessed by fit statistics: AIC and BIC) had only confidence, dataset, and their interaction as fixed-effect predictors, see Figure \ref{fig:Exp2_accmodel}. 

The best fit model indicates confidence calibration, again, accounts for a moderate amount of variance in fixed and varying effects, conditional Pseudo-$R^2$ = 0.13. Including point vs. distribution or visualization type did not appear to improve model fit or to predict accuracy performance above and beyond what is predicted by dataset and confidence. See data and code links in the supplementary material for details of the model.

Thus, contrary to the pre-registration, neither our accuracy or confidence calibration results indicate support for our Experiment 2 hypotheses. The hypotheses were that performance would be better with distribution information for UQ than for point-estimated UQ, and that within the distribution condition, performance would be better with dotplots than with the needle visualization.  

Additionally, we analyzed RT in Experiment 2 (see right side of Figure \ref{fig:all_rt_results}) to rule out speed/accuracy tradeoffs in the performance results. An omnibus 2 x 2 x 3 ANOVA indicated no significant effect of dataset ($F(2,588) = 0.09, {p} = 0.92, \eta_p^2 < 0.001$), point vs. distribution ($F(1,588) = 0.06, {p} = 0.81, \eta_p^2 < 0.001$), or visualization type ($F(1,588) = 0.01, {p} = 0.91, \eta_p^2 < 0.001$) upon RT. Neither was there evidence of significant interaction effects among the three manipulated variables.

\subsection{Exploratory Analyses}
\label{sec:expa}
Although this work was not specifically designed to assess whether the combination of humans plus AI generally exceeded the accuracy of AI predictions, we conducted exploratory analyses to investigate how often this occurred. In both experiments, we found that a small subset of participants were able to outperform the accuracy of the AI predictions they received (see Table \ref{table:outperformAI}). AI uncertainty enabled humans to outperform the AI accuracy more frequently and with patterns of mean differences suggesting greater improvements. 

\begin{table}[h]
\caption{Number of participants who outperformed the AI}
\label{table:outperformAI} 
\vskip 0.15in
\begin{center}
\begin{small}
\begin{tabular}{l r r r} 
\toprule
Human Accuracy & \textgreater AI (\textit{n}) & $\le$AI (\textit{n}) &\textgreater AI (\%)\\  
 \midrule
 Experiment 1\\ 
 \hspace{3mm}AI Prediction & 2 & 148 & 1.33\%\\
 \hspace{3mm}AI Uncertainty & 7 & 141 & 4.73\%\\
 Experiment 2 \\ 
 \hspace{3mm}Point Needle & 8 & 142 & 5.33\%\\
 \hspace{3mm}Point Dotplot & 8 & 142 & 5.33\%\\
 \hspace{3mm}Dist Needle & 11 & 139 & 7.33\%\\
 \hspace{3mm}Dist Dotplot & 7 & 143 & 4.67\%\\
\bottomrule
\end{tabular}
\end{small}
\end{center}
\end{table}

\section{General Discussion, Limitations, and Future Work}
\label{sec:gendisc}
The overall results of Experiment 1 showed that providing AI UQ to human decision-makers improved accuracy and confidence calibration performance over and above providing an AI prediction alone. In Experiment 2, we did not find meaningful differences between different representations of UQ. Taken together, our findings suggest that the \textit{benefit of AI UQ may not be overly sensitive to the representation of the UQ information}. Also, adding more information (here, the UQ distribution), even though task-relevant, did not improve decision-making which is consistent with prior research~\cite{marusich_info, gigerenzer2009homo, Alufaisan_Marusich_Bakdash_Zhou_Kantarcioglu_2021}. 

Here, the humans and AI had identical information so we could evaluate well-calibrated UQ. Normatively, algorithms tend to outperform human decision-making in a variety of tasks; a clear exception is when people have knowledge the algorithm does not \cite{dawes1989clinical}. In such situations, it is possible that the AI and human combined can produce better performance than either alone \cite{cummings2014man}. Exploratory results suggested AI uncertainty may increase the frequency of people exceeding AI accuracy, although this was not a common occurrence; likely because both the human and the AI had the same information.


In this work, we did not compare different UQ techniques but did demonstrate that, using a Brier score, our UQ technique performed well in practice. It is possible that UQ techniques that do not perform as well as ours may not improve human decision-making accuracy. In future work, we plan to use other UQ techniques, including less effective ones to understand the impact of UQ quality on human decision-making accuracy. Similarly, behavioral experiments were limited to comparing point versus distributions and two visualizations of predictive uncertainty. It is possible that other visualizations may make providing distribution information more effective, and we plan to conduct future work assessing more UQ visualization techniques. 

Finally, we used a relatively simple binary classification decision-making task. In more complex application domains, where humans encounter multi-class classification problems, the impact of UQ information on human decision making could be different. Future work should explore multi-class classification problems, as well as the use of AI UQ for complex tasks where people, such as experts, have knowledge unknown to the AI model. 

\section{Conclusion}
\label{sec:conc}


Our extensive behavioral experiments show that providing \textit{high quality} AI uncertainty information improves human decision accuracy and confidence calibration over the AI prediction alone. This human performance benefit was not limited to only a specific visual representation of UQ information. In previous work, there is an absence of evaluating calibration for AI UQ \cite{lai2021towards}. Here, we used Brier scores to quantify the calibration of our implementation of AI UQ. We showed the AI UQ used here was well-calibrated by leveraging the existing class labels of test instances.  

\section{Impact Statements}
Recently, understanding how humans and AI systems can work better together has emerged as an important challenge. Although previous work has explored how and when explainable AI may help human decision making, the impact of providing uncertainty information to humans has not been explored in depth in the context of AI systems. To address this challenge, in this work, we explore whether providing uncertainty information to humans may improve decision making and potentially correct errors caused by the AI models. 
\section*{Software and Data}
See supplementary material or \url{https://osf.io/cb762/}. 

\section*{Acknowledgements}
We thank Mary Grace Kozuch for helpful feedback on a previous version of this paper. 

The views and conclusions contained in this document are those of the authors and should not be interpreted as representing the official policies, either expressed or implied of the U.S. DEVCOM Army Research Laboratory or the U.S. Government. The U.S. Government is authorized to reproduce and distribute reprints for Government purposes notwithstanding any copyright notation. M.K. and Y.Z. were supported in part by NSF awards OAC-1828467, DMS-1925346, CNS2029661, OAC-2115094, ARO award W911NF-17-1-0356, and a gift from Cisco Inc.



\bibliography{icml2024}
\bibliographystyle{icml2024}




\end{document}